\documentclass[10pt,conference]{IEEEtran}
\IEEEoverridecommandlockouts
\usepackage{cite}
\usepackage{amsmath,amssymb,amsfonts}
\usepackage{algorithmic}
\usepackage{graphicx}
\usepackage{textcomp}
\usepackage{xcolor}
\def\BibTeX{{\rm B\kern-.05em{\sc i\kern-.025em b}\kern-.08em
		T\kern-.1667em\lower.7ex\hbox{E}\kern-.125emX}}
\usepackage{multirow}

\makeatletter
\newcommand{\linebreakand}{%
  \end{@IEEEauthorhalign}
  \hfill\mbox{}\par
  \mbox{}\hfill\begin{@IEEEauthorhalign}
}
\makeatother

\begin{document}
\title{Convolutional Neural Networks for Time-dependent Classification of Variable-length Time Series}

\author{\IEEEauthorblockN{Azusa Sawada}
\IEEEauthorblockA{
	\textit{NEC Corporation}\\
	Kawasaki, Japan \\
	swd02@nec.com}
\and
\IEEEauthorblockN{Taiki Miyagawa}
\IEEEauthorblockA{
	\textit{NEC Corporation}\\
	Kawasaki, Japan \\
	miyagawataik@nec.com}
\and
\IEEEauthorblockN{Akinori F. Ebihara}
\IEEEauthorblockA{
	\textit{NEC Corporation}\\
	Kawasaki, Japan \\
	aebihara@nec.com}
\linebreakand
\IEEEauthorblockN{Shoji Yachida}
\IEEEauthorblockA{
	\textit{NEC Corporation}\\
	Kawasaki, Japan \\
	s-yachida@nec.com}
\and
\IEEEauthorblockN{Toshinori Hosoi}
\IEEEauthorblockA{
	\textit{NEC Corporation}\\
	Kawasaki, Japan \\
	t.hosoi@nec.com}
}
\maketitle

\begin{abstract}
Time series data are often obtained only within a limited time range due to interruptions during observation process. To classify such partial time series, we need to account for 1) the variable-length data drawn from 2) different timestamps. To address the first problem, existing convolutional neural networks use global pooling after convolutional layers to cancel the length differences. This architecture suffers from the trade-off between incorporating entire temporal correlations in long data and avoiding feature collapse for short data. 
To resolve this trade-off, we propose Adaptive Multi-scale Pooling, which aggregates features from an adaptive number of layers, i.e., only the first few layers for short data and more layers for long data.
Furthermore, to address the second problem, we introduce Temporal Encoding, which embeds the observation timestamps into the intermediate features. Experiments on our private dataset and the UCR/UEA time series archive show that our modules improve classification accuracy especially on short data obtained as partial time series.

\end{abstract}


\section{Introduction}\label{sec:intro}
\begin{figure}[t]
	\begin{center}
		\includegraphics[width=8.6cm,trim=25 160 210 80,clip]{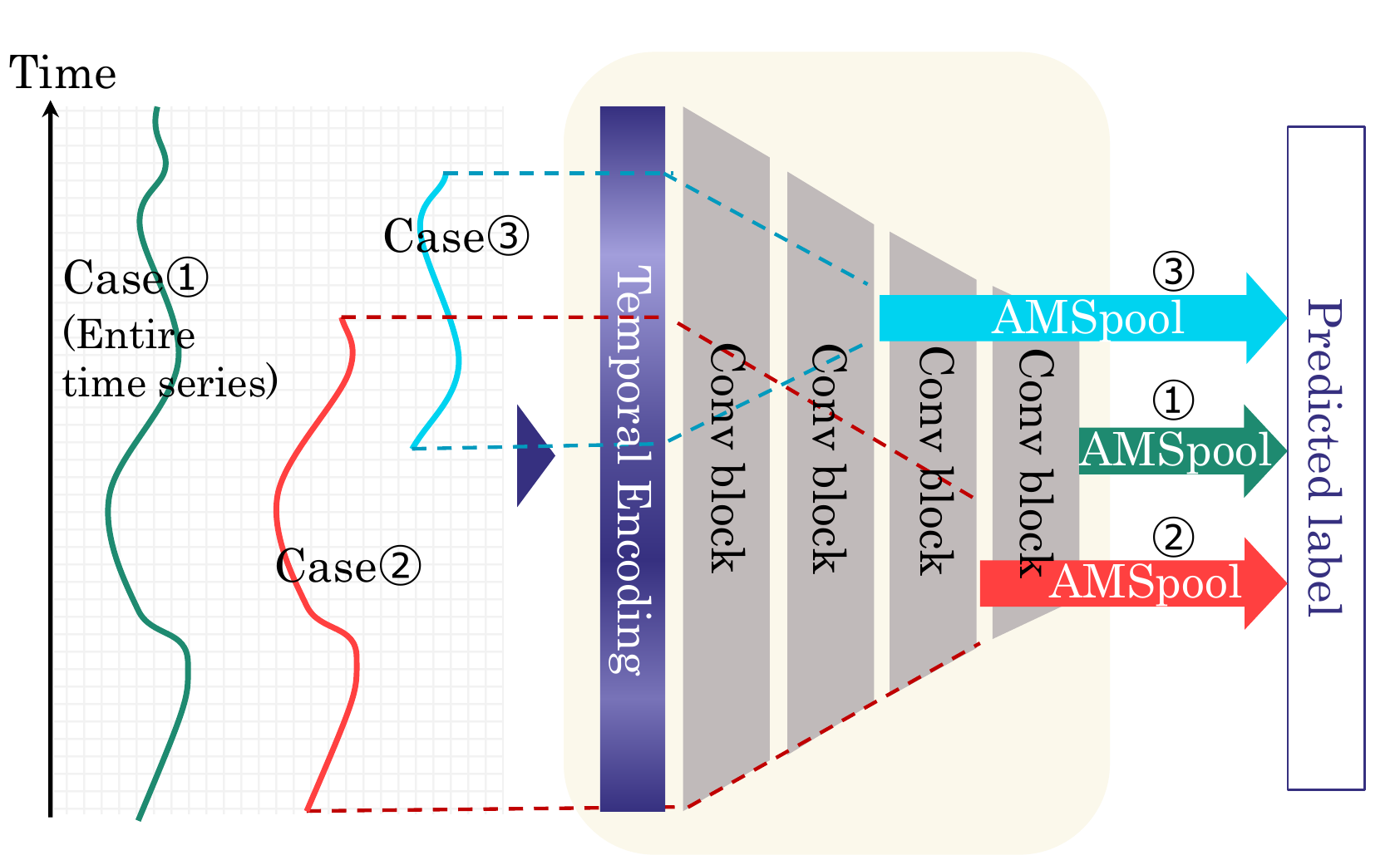}
	\end{center}
	\caption{pTSC and proposed method. 
	The three cases of inputs on the left have different lengths and timestamps.
	Our method makes it possible to classify all of them with a single CNN model.
	Each prediction flow is indicated by different colors (case 1: green, case 2: red, case 3: light blue).
	First, the inputs are concatenated with the corresponding parts of Temporal Encoding (TE) to embed when inputs are observed. 
	Next, they are passed to convolutional blocks, where the length shrinks after each block as shown by dashed lines.
	AMSpool outputs the final features for label prediction, using only the convolutional features truncated at intermediate layers. The truncation is done at the block immediately before the extrapolated dashed lines cross. See also Fig. \ref{fig:RF} for the definition of the dashed lines.
	}
	\label{fig:concept}
\end{figure}

\renewcommand{\thefootnote}{}
\footnote[0]{©2022 IEEE.  Personal use of this material is permitted.  Permission from IEEE must be obtained for all other uses, in any current or future media, including reprinting/republishing this material for advertising or promotional purposes, creating new collective works, for resale or redistribution to servers or lists, or reuse of any copyrighted component of this work in other works.}
\renewcommand{\thefootnote}{\arabic{footnote}}
We define a novel classification problem of time series, where the input data are given only partially in limited time ranges (Fig.~\ref{fig:concept} left), which we call \textit{partial time series classification} (pTSC).
pTSC appears, for example, in detecting foreign matters apart from air bubbles in liquid drugs such as vaccines. The trajectory patterns of the foreign matters and air bubbles contain discriminative information that enables accurate classification; however, the trajectories are often fragmented because of the tracking failures and occlusions. 
Note that pTSC is \textit{not} equivalent to variable-length TSC, where entire time series are given as inputs.
In pTSC, on the other hand, inputs are given as parts of the entire time series. 

\begin{figure}[t]
	\begin{center}
		\includegraphics[width=8.5cm,trim=0 205 270 80,clip]{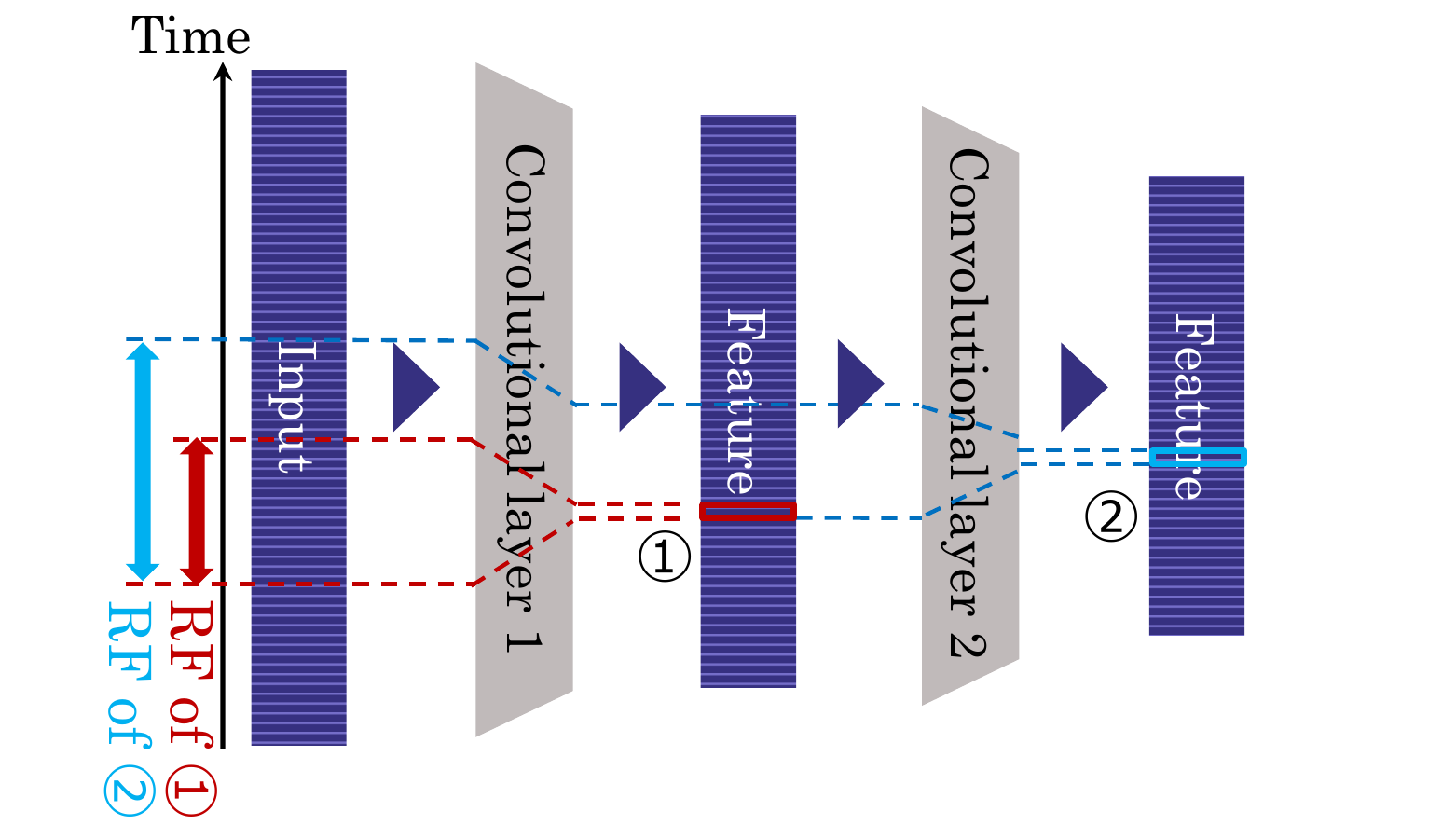}
	\end{center}
	\caption{
	Receptive field (RF) of feature elements in convolutional layers.
    The RF of a given feature element (1: red and 2: blue box) is defined as the region of the input series (vertical red and blue arrow, respectively) used to calculate the feature element. 
    The width of the broken lines narrow down according to the kernel size of the convolutional layer.
	}
	\label{fig:RF}
\end{figure}

In this paper, we address two problems in pTSC: 1) variable-length inputs obtained as 2) the fragments drawn from different timestamps.
The first problem is closely related to the \textit{receptive field} (RF) of a given feature element, which is defined as the region of the input series used to calculate the feature element (Fig.~\ref{fig:RF}). In previous works, convolutional neural networks (CNNs) with global pooling layers are often used to cancel the variation of length after convolutional layers~\cite{xception,ShapeNet,VLtraining}.
Because global pooling layers ignore the order of their inputs, CNNs cannot fully capture temporal correlations longer than the RFs of the features before the global pooling layers. Thus, small RFs may potentially be detrimental in classifying long time series.
On the other hand, the features cannot be computed if the inputs are shorter than the RFs of the features; in this case, padding to the inputs is often used but it contaminates the feature. Then, long RFs may cause degradation of classification accuracy on short time series.
Therefore, CNNs suffer from a trade-off between the long and short series.
A na\"ive solution is to train multiple classifiers independently for all different lengths, but it leads to a large training cost. 
Moreover, as mentioned above as the second problem in pTSC, we have to handle the partial time series at different timestamps that are parts of the original time series. Therefore, it is desirable that the classification model knows the original timestamps of the fragments, which may contribute to improving the classification accuracy.


To address these two problems, we propose Adaptive Multi-scale Pooling (AMSpool) and Temporal Encoding (TE). AMSpool is an architecture that aggregates features from an adaptive number of layers (the three right-most arrows in green, red, and blue in Fig. \ref{fig:concept}). The number of layers is controlled to adjust the RF size of the feature to the input size. Therefore, AMSpool enables us to classify short data without the contamination with padding, while keeping large RF sizes for long data: resolving the aforementioned trade-off. TE concatenates a part of long vectors to the input series to embed the observation timestamps (Fig. \ref{fig:concept}). Thus, TE addresses the second problem mentioned above.
These two modules can be introduced to any architecture consisting of multiple convolutional layers.


The experiment on our private dataset for trajectory classification shows that our modules improve accuracy, e.g., from $84.8\%$ to $89.5\%$, on short data from baseline models.
In addition, to conduct experiment on a variety of datasets from different domains, we use the UCR/UEA archive~\cite{UEA2018,UCR2019} and show that AMSpool works well on the datasets with the input length comparable to the RF of the last convolutional feature.

Our contribution is threefold:
\begin{enumerate}
    \item We define and tackle a novel challenging classification problem, pTSC. 
    \item We propose a novel architecture, AMSpool, to address the aforementioned trade-off between long and short data. 
    \item We propose TE, which embeds the observation timestamps into the input series and thus is expected to contribute to classification accuracy.
\end{enumerate}

\section{Related Work}\label{sec:related}
Time series classification (TSC) is an important task in analyzing various data such as motion, sound, and electrical signals. General TSC models have been build upon the UEA/UCR archive~\cite{UEA2018,UCR2019}, which contains diverse data types~\cite{Bakeoff2017,review2019}. 1D-CNNs are well-performing models on the archive. FCN\cite{strongBaseline2016} and ResNet\cite{strongBaseline2016} were the early attempts at end-to-end deep learning models. ResNet was shown to be more accurate than other models~\cite{review2019}. Recently InceptionTime\cite{inception2020} has shown comparable accuracy to the transformation-based TSC method~\cite{hive-cote} which is difficult to use for a large number of data or long series. These 1D-CNNs consist of convolutional layers, batch normalization layers, global average pooling layers, and skip connections. The Recurrent Neural Network (RNN) is another well-known type of models for sequential inputs but no RNN-based models perform as well as existing CNNs~\cite{RNNcomp,review2019}.

pTSC can be regarded as one of the classification problem of time series with missing values. However, in pTSC, the missing values appear as consecutive regions and often last much longer than observed regions, which makes it difficult to fill the missing values~\cite{imputation2020}. To the best of our knowledge, there has not been a study that directly address the problem equivalent to pTSC, where both variable length and variable time offset are included. In the existing work on liquid inspection, trajectory data are simply classified using handcraft features without accounting for timestamp differences~\cite{liquid2020}.

\subsection{Variable Length}
The variable-length problem was often avoided in deep learning for general TSC because some time series datasets can be classified as fixed-length data, with preprocessing such as interpolation to adjust length~\cite{review2019}. However, in pTSC, where the input length changes due to an external reason such as an interruption during observation, preprocessing causes severe inconsistency in effective sampling rates to obtain fixed-length data.

In existing CNN models, pooling layers are utilized to cancel differences in series length. Global pooling was proposed as the parameter-free alternative of fully-connected layers, but it enables variable-length and fully-convolutional architecture~\cite{maskedGP}. ShapeNet~\cite{ShapeNet} includes global max pooling in feature extraction CNNs to map variable-length subsequences in a common feature space. XceptionTime~\cite{xception} uses adaptive average pooling after convolutional layers to obtain fixed features with fixed number of components in time axis. Dynamic time pooling~\cite{DTP2021}, which determines pooling segments adaptively, was proposed to extend fixed pooling segments. Although these single pooling modules can summarize feature series in a global or segment-wise manner, the length is limited because a pooling segment must have a positive size. To design CNNs within this limitation, the RF of convolutional blocks cannot be larger than the minimum input length, which makes it impossible to capture long-term correlations. 

The selection of RF size or kernel sizes has been studied to extract features of suitable temporal scales. There are methods for sample-wise selection by predicting mask of kernels~\cite{DynamicMSCNN} and by using selective kernel layer~\cite{SKNet2019} that apply attention to outputs from two convolutional layers with different kernel size~\cite{VLtraining}. Omni-scale CNN~\cite{OS-CNN} (OS-CNN) was proposed to address dataset-wise selection. It is designed to contain all RF sizes as the combinations of kernel sizes in two layers. These models select a more suitable RF or cover all possible RFs in feature extraction, but the maximum RF is still limited by the minimum length at the pooling layers.

We propose a module that can realize maximum RF larger than the minimum input length so that it can be applied to a wide range of length without disregarding long temporal correlation.




\subsection{Variable Timestamps}
Positional encoding that adds position-dependent variables to the input~\cite{AbsolutePE} is used to feed positional information to CNNs for sequential input and output. Positional encoding can be found in transformers~\cite{Transformer} to encode relative position rather than absolute one because they use self-attention which is position-agnostic. In this paper, we utilize absolute positional encoding to represent the differences in input timestamps by concatenating time-dependent variables to the input. 


\section{Method}\label{sec:method}
\begin{figure*}[t]
 	\begin{center}
    	\begin{minipage}{.64\linewidth}
    	    \textbf{(a)}\\
        	\includegraphics[width=11.5cm,trim=10 0 0 70,clip]{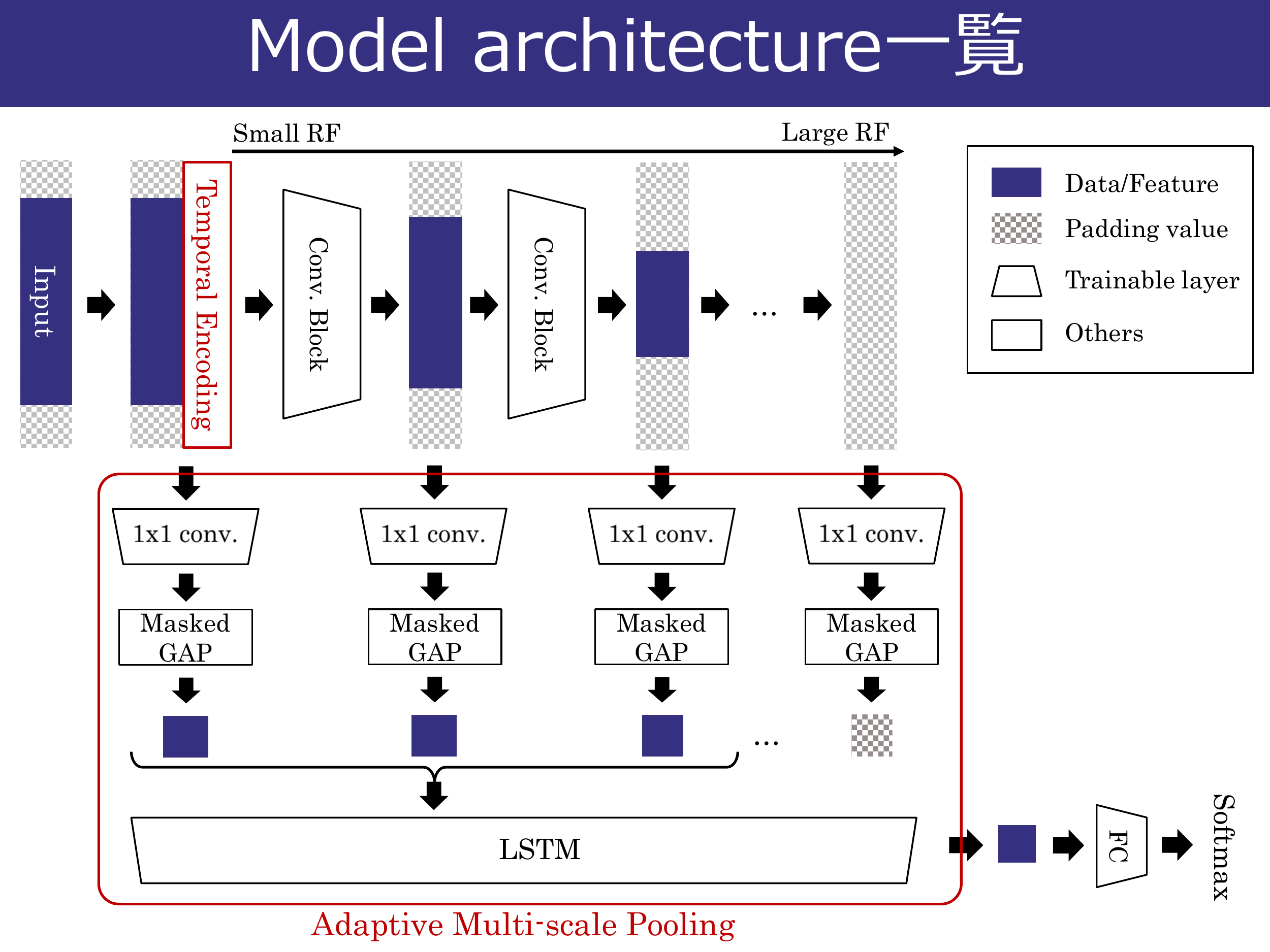}
        \end{minipage}
        \begin{minipage}{.34\linewidth}
    		\textbf{(b)}\\
    		\includegraphics[width=6.5cm,trim=20 10 140 240,clip]{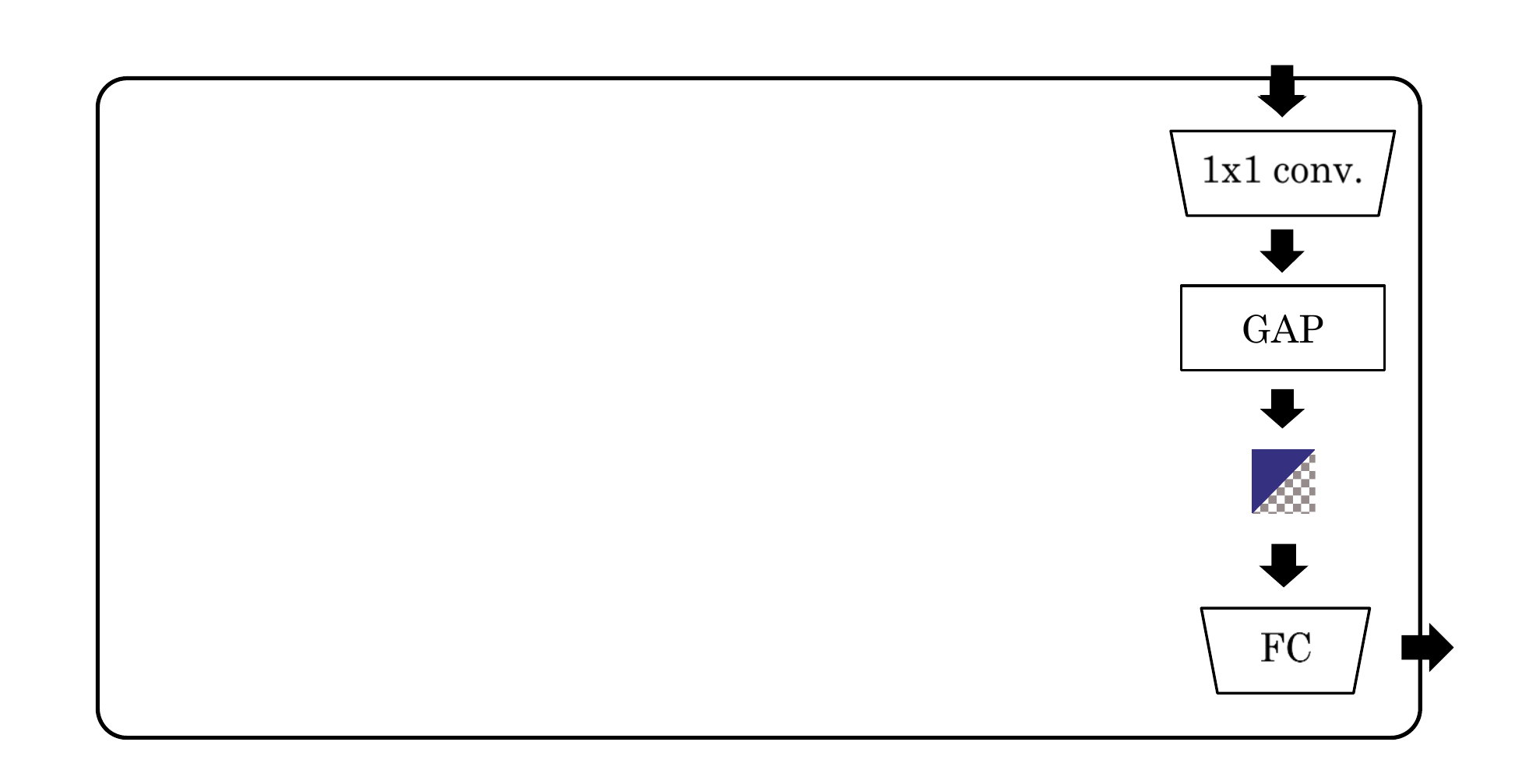}\\
    		\textbf{(c)}\\
    		\includegraphics[width=6.5cm,trim=20 10 140 240,clip]{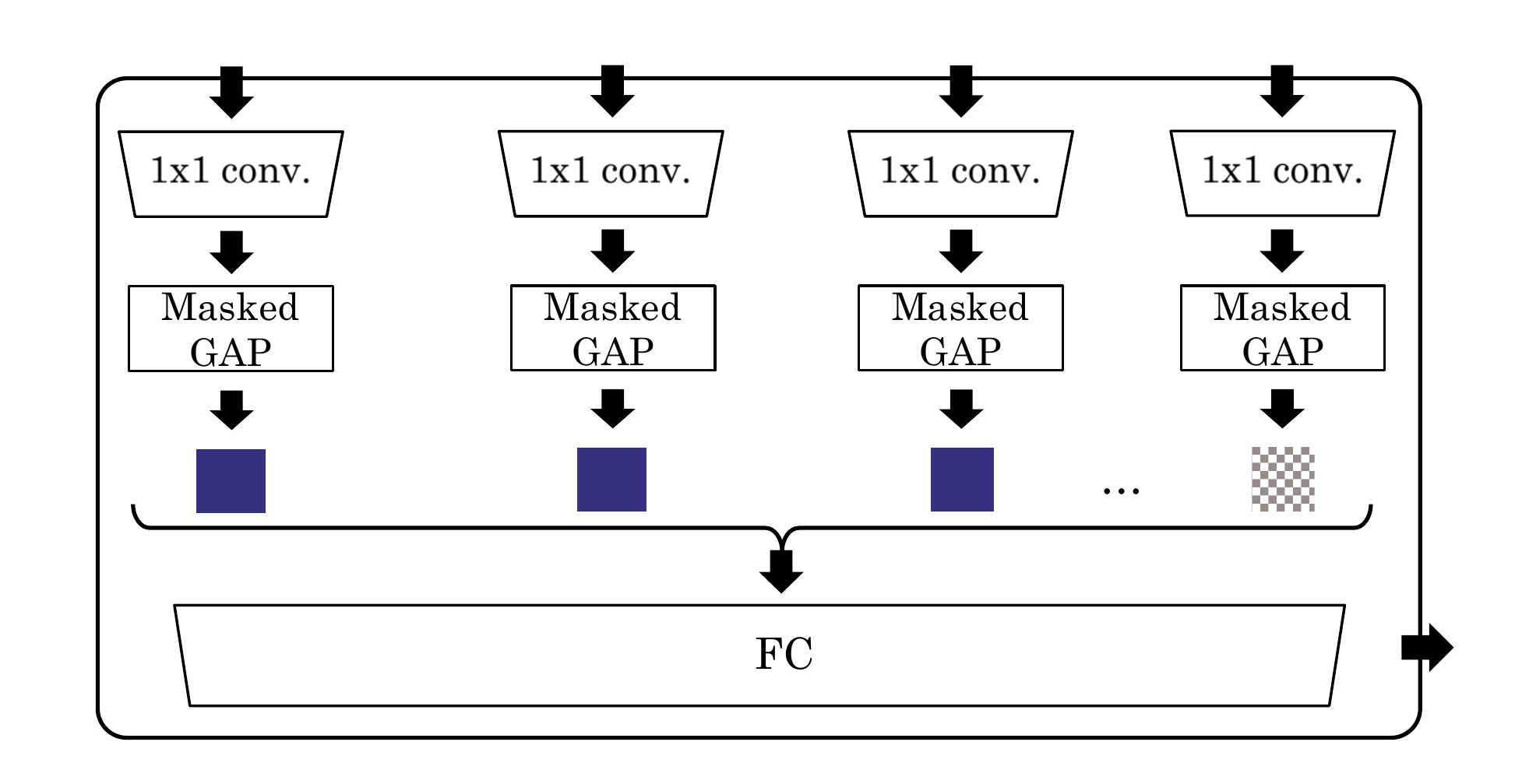}\\
    		\textbf{(d)}\\
    		\includegraphics[width=6.5cm,trim=20 10 140 240,clip]{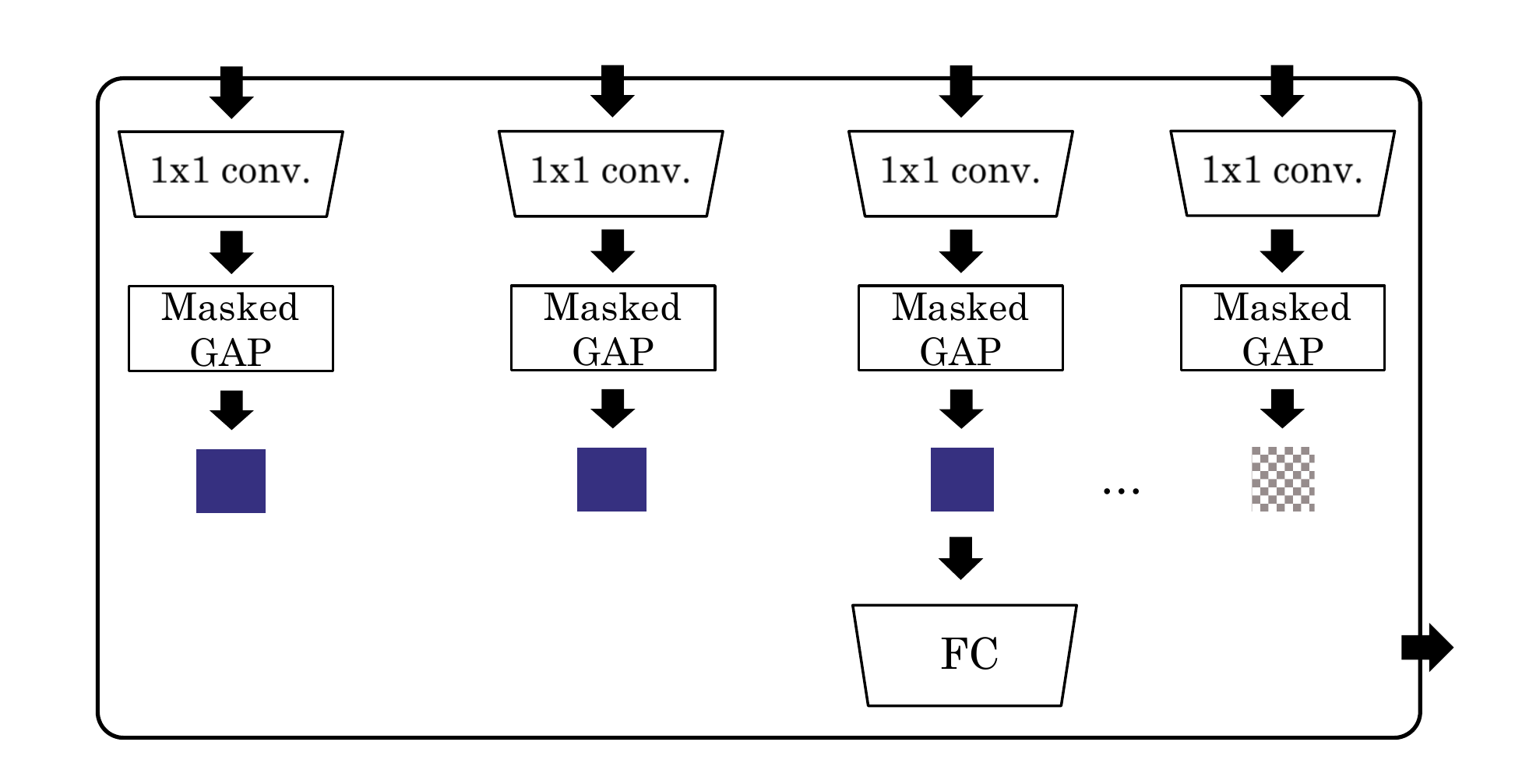}
		\end{minipage}
	\end{center}
	\caption{Overview of proposed architecture and variants. (a) In the proposed architecture, Temporal Encoding is concatenated before convolutional layers. Feature maps from all convolutional blocks are fed to pointwise convolution and GAP layer. The outputs are concatenated as a variable-length feature series from small to large RF, truncated before exceeding input length. Adaptive Multi-scale Pooling extracts fixed-size feature from the sequence. (b)-(d) the alternative variants of Adaptive Multi-scale Pooling in (a), corresponding to BaseCNN, MSCNN and ASCNN used in our ablation study.}
	\label{fig:model}
\end{figure*}

Our method consists of two modules for 1D-CNN, Adaptive Multi-scale Pooling (AMSpool) and Temporal Encoding. 
The overview of the model with the modules is shown in Fig.~\ref{fig:model}a. We explain the detail of the modules in this section.

Let us clarify the notation of our problem. TSC is a task for predicting class label $y \in {1,...,N}$ for a given time series $\textbf{x} \in \mathcal{R}^{D \times T}$, where $D$ is the number of variables and $T$ is the length in time. In our problem, both the initial timestamp $t_1$ and length $T$ vary by $\textbf{x}$. We assume $T$ ranges from $T_{min}$ to $T_{max}$ and there are timestamps from $1$ to $T_{max}$ and time interval is $1$.

\subsection{Temporal Encoding}\label{subsec:te}
As mentioned in Sec.~\ref{sec:intro}, partial time series have different ranges of timestamps. \textit{Temporal Encoding} (TE) is the component to encode timestamps of inputs before fully-convolutional neural networks. It is concatenated to the input time series frame-by-frame at the corresponding time range as additional channels (Eq.~\ref{eqn:te}). The size of temporal encoding is set to the same as the input size in our experiments.
\begin{equation}\label{eqn:te}
(\textbf{f}_0)_i = concat(x_i, e_{t_i}),~~i=1,...,T
\end{equation}
TE is similar to positional encoding (or positional embedding) in natural language processing~\cite{AbsolutePE}. They encode the position information of inputs and outputs to fully convolutional networks for sequence-to-sequence tasks\footnote{Positional encoding is also used to introduce information on relative positions between words into self-attention modules.}. In our case, it is used to represent not only intra-series positions but also inter-series difference in initial positions in the entire time range.

Here, concatenation form is utilized for TE rather than addition form which is used in original positional encoding~\cite{AbsolutePE}. This is because we consider timestamp information as an additional input besides given partial time series. There are also another reason that the additional form of positional encoding may make it difficult for entire models to converge when it is added to raw time series data as indicated in the case of modeling acoustic sequences by self-attention models~\cite{AcousticSA}. The concatenated temporal encoding and corresponding kernels in the first convolutional layer are put together into the bias term of the layer without any difference, but we treat temporal encoding as a module separated from successive layers.

In the implementation, all time series are padded to a fixed length while preserving temporal position. When a given time series starts at $t=T_0$ and ends at $T_0+T-1$, for example, it is processed pre-padding by $T_0-1$ frames and post-padding by $T_{max}-T_0-T+1$ frames. Then, the time series are concatenated with TE at corresponding timestamps without any sorting components of TE. The padding effect is masked out later at global pooling layers. Note that temporal encodings can easily be applied when the conditions are cyclic in time.

\subsection{Adaptive Multi-scale Pooling}
To capture the entire temporal correlation of length $T_max$, a model needs to be designed to have an RF comparable to $T_max$. The existing CNNs cannot have an RF larger than $T_{min}$ without any preprocessing to compensate for lack of input length. 

To design a CNN architecture that can realize RF about $T_{max}$ at maximum and to control RF down to $T_{min}$, we introduce a mechanism to truncate convolutional features with larger RF than the input length. Let us denote the number of convolutional blocks of the base CNN as $L$ and the output of $l$-th block as $\textbf{f}_l$.

Point-wise convolution $F_l$ is performed on feature maps $\textbf{f}_l$ and optionally input layer $\textbf{f}_0$ to project them to the same number of channels. Then the outputs are passed to global average pooling (GAP) layers as shown in Fig.~\ref{fig:model}a. 
\begin{equation}\label{eqn:reshape}
\textbf{z}_l = GAP(F_l(\textbf{f}_l))
\end{equation}
Note that $\textbf{z}_l$ can be empty for $l$ s.t. RF of the $l$-th block is larger than $T$.
Then $\textbf{z}_l$s in ascending order of $l$ can be seen as a feature series $\textbf{z}$ with length $T'$, where $T'$ equals the number of convolutional blocks whose RFs do not exceed $T$. This $\textbf{z}$ contains features of multiple RFs, namely multi-scale features, with the maximum RF controlled.

Because $\textbf{z}$ is ordered from smallest $l$ and ends at variable length, we apply a recurrent module LSTM~\cite{lstm} to aggregate multi-scale features to obtain a fixed-length feature. A CNN with this module remains fully-convolutional in time axis.
\begin{equation}\label{eqn:lstm}
\textbf{z}_{f} = LSTM(\textbf{z})
\end{equation}
$\textbf{z}_{f}$ is passed to output layer with softmax function through a fully-connected layer which has the output dimension $N$.

The implementation of models with AMSpool shares common modules with CNNs for fixed-length inputs when padding is applied to inputs. The difference is masking in GAP layers and batch normalization layers~\cite{BN2015} (if any). The model tracks the valid region that is purely calculated from the input data on the time axis after convolutional blocks. The outside of this region is masked at GAP layers and batch normalization layers in order to eliminate the padding effect. For example, when the input data is placed in $t=T_0,...T_0+T-1$, the valid region in the feature map after convolution with kernel size $2k+1$ and padding size $k$ for both sides is $t=T_0+k,...T-T_0-k-1$.




\section{Experiments}\label{sec:exp}
We conducted two sets of experiments.
The first set was the experiments on our private dataset for trajectory classification which show a typical pTSC setting. We compared our models with existing TSC models with preprocessings and conducted ablation study of our modules.
The second set was the experiments using public TSC datasets to explore our modules with an existing TSC model as the base CNN structure.

\subsection{Trajectory dataset}\label{sec:exp2}

\begin{figure}[t]
	\begin{center}
		\includegraphics[width=8.5cm,trim=20 0 30 30,clip]{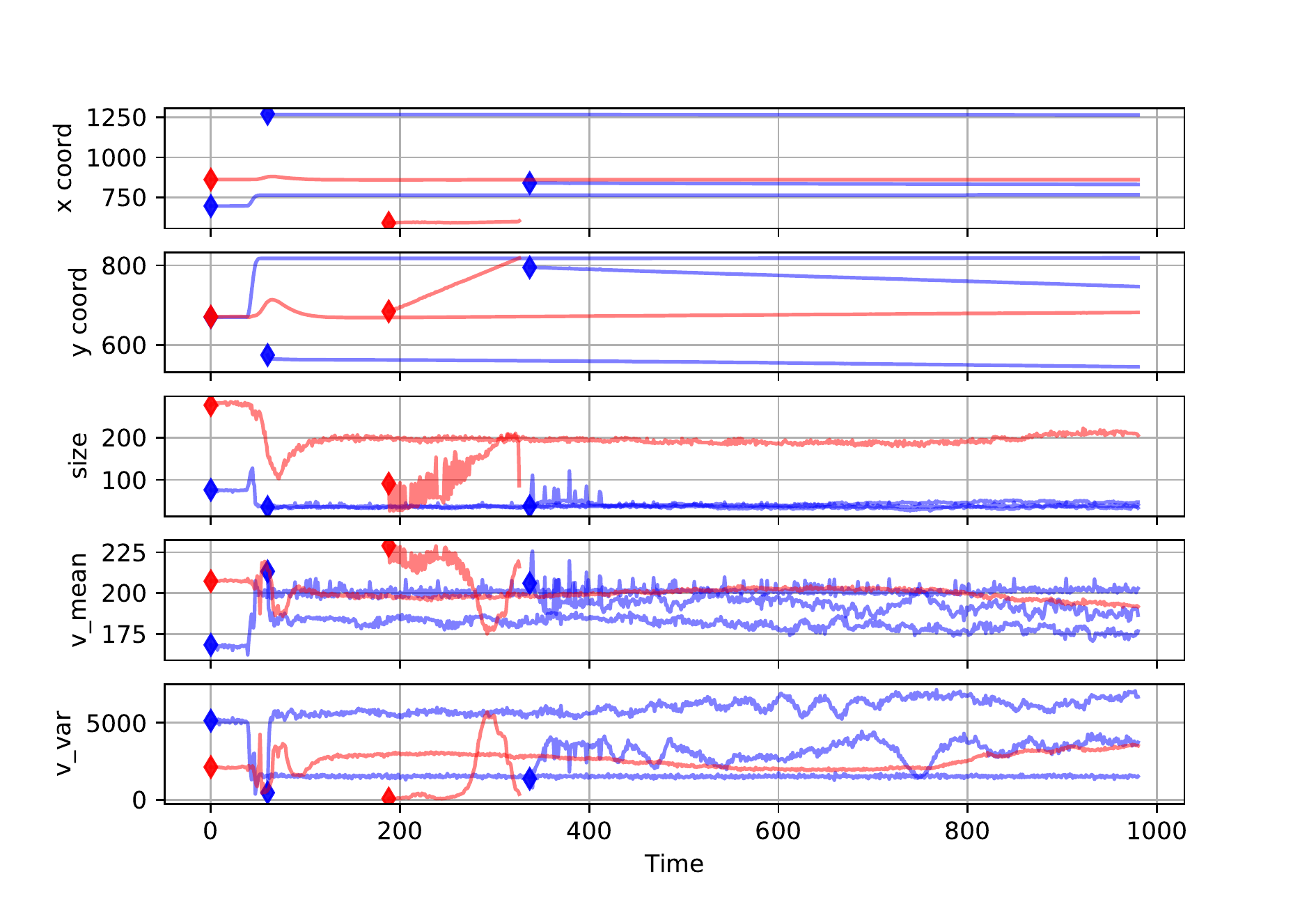}
	\end{center}
	\caption{Example plot of five components of trajectory data. Red curves indicate anomaly class and blue curves are normal class including bubbles and scratches. Diamond markers show initial components.}
	\label{fig:tracklet_example}
\end{figure}

Trajectory data are tracks of small particles in transparent liquid containers. The classes are defined as ``anomaly" for foreign particles and ``normal" for bubbles, scratches, and noise. Each trajectory is generated by object tracking of videos and consists of five components: horizontal coordinate (x coord.), vertical coordinate (y coord.), size, mean brightness (v mean) and variance of brightness (v std) as shown in Fig. \ref{fig:tracklet_example} (all components are normalized into $[0,1]$ when they are input to models). All videos starts at common timestamp $t=1$ and the containers rotate at $t=40$ to induce the movement of particles. The length is in $[80,980]$ after removing too short data that contain almost only noise. The test set is separated so that it does not include the data extracted from common videos in the training set.

The base structure for our proposals is designed to have an RF comparable to the maximum input length. It has six convolutional layers with kernel size $7, 5, 5, 3, 3, 3$, each followed by batch normalization layers and ReLU activation with max pooling layers with kernel size $2, 4, 2, 4, 2, 4$ inserted, respectively.
BaseCNN is the model that has the base structure followed by GAP and two fully-connected layers (Fig.~\ref{fig:model}b), and inputs are fixed-length data after zero-padding\footnote{In this experiment, zero-padding is performed in the same way used in the implementation of TE explained in Subsec.~\ref{subsec:te}}. AMSCNN refers to the model with AMSpool and TE, which shares the convolutional blocks and final fully-connected layers with BaseCNN.

All models are trained with Adam optimizer~\cite{adam} for $100$ epochs using random crop to minimize crossentropy loss weighted by the inverse class frequency. We use $20~\%$ of the training samples as a validation set to evaluate AUROC (Area Under the Receiver operating Characteristic curve) for tuning hyperparameters: initial learning rate, mini-batch size, weight decay and epoch to multiply learning rate by $0.1$. We report balanced accuracy for three group of test set devided by data length as the average of twenty runs with different random seeds.

Table~\ref{tab:syringe3_result} compares the results of AMSCNN-te with BaseCNN and existing models with preprocessings (``pad" and ``intp" refers to zero-padding and interpolation, respectively). On short and middle data, AMSCNN-te improves the balanced accuracy by $4.7 \%$ and $1.1 \%$ compared with BaseCNN. The improvement on short data is confirmed to be statistically significant (with a p-value of $0.001$) using Steel-Dwass method, which is the non-parametric test for comparison between all pairs. Existing TSC models with preprocessings show much lower accuracies in short and middle data. This suggests that the classification accuracy on short data degrages due to the severe distortion of the data by preprocessing, which is avoided in our AMSCNN-te.  Conversely, on long data, AMSCNN-te shows lower balanced accuracy than BaseCNN and OS-CNN, although this difference is smaller than improvements in short and middle data. 

\begin{table}[t]
	\caption{Results on trajectory dataset. Values are balanced accuracies in three data groups divided by series length (short~/~middle~/~long). All values are the average of 20 runs. Numbers in parentheses represent standard deviation. Highest average value in each column is in bold font.}
	\label{tab:syringe3_result}
	\begin{center}
		\begin{tabular}{lccc}
		    \hline
			\multicolumn{1}{c}{\bf Variant} & \multicolumn{3}{c}{\bf Balanced accuracy} \\
			& Short & Middle & Long\\
			\hline \hline
			InceptionNet\cite{inception2020}-pad & 0.582~(0.058) & 0.641~(0.049) & 0.696~(0.090)\\
            InceptionNet\cite{inception2020}-intp & 0.619~(0.067) & 0.691~(0.092) & 0.733~(0.123)\\
            OS-CNN\cite{OS-CNN}-pad & 0.824~(0.018) & 0.862~(0.022) & 0.862~(0.013)\\
            OS-CNN\cite{OS-CNN}-intp & 0.835~(0.016) & 0.855~(0.017) & \textbf{0.903}~(0.014)\\
            \hline
            BaseCNN & 0.848~(0.012) & 0.917~(0.016) & 0.888~(0.014)\\
            AMSCNN-te~(ours) & \textbf{0.895}~(0.015) & \textbf{0.928}~(0.014) & 0.867~(0.019)\\
            \hline
		\end{tabular}
	\end{center}
\end{table}

Next, we conduct the ablation study using models in Table~\ref{tab:methods_model}. \textit{TE} column refer to the existence of TE. \textit{Adaptive RF} indicates whether RF size is adjusted according to the input length, and \textit{multi-scale feature} indicates whether feature contains multiple RFs (in classifying a single input). MSCNN is the model which concatenates features from multiple blocks after global average pooling (Fig.~\ref{fig:model}c), which can be seen as multi-scale statistics pooling~\cite{VLtraining} modified to exclude standard deviation features, instead of GAP in BaseCNN. ASCNN is a structure that excludes multi-scale components from AMSCNN (Fig.~\ref{fig:model}d).

\begin{table}[t]
	\caption{Model list of architectures in ablation study.}
	\label{tab:methods_model}
	\begin{center}
		\begin{tabular}{lccccc}
			{\bf Variant} & \textbf{TE} & \textbf{Adaptive RF} & \textbf{Multi-scale feature} \\
			\hline
			BaseCNN & - & - & - \\
			BaseCNN-te & \checkmark & - & - \\
			MSCNN & - & - & \checkmark \\
			MSCNN-te & \checkmark & - & \checkmark \\
			ASCNN & - & \checkmark & - \\
			ASCNN-te & \checkmark & \checkmark & - \\
			AMSCNN & - & \checkmark & \checkmark \\
			AMSCNN-te & \checkmark & \checkmark & \checkmark \\\
		\end{tabular}
	\end{center}
\end{table}

Table~\ref{tab:syringe3_result_all} shows the results of the ablation study. 
Between the models with and without TE, the average accuracy for short series tends to be improved with TE (except for ASCNN). This suggests that models with TE can account for the time dependency which becomes critical when given time series are short parts.

\begin{table}[t]
	\caption{Ablation study on trajectory dataset. Values are balanced accuracies on three data groups divided by series length (short~/~middle~/~long). Numbers in parentheses represent standard deviation. Highest average value in each column is in bold font.}
	\label{tab:syringe3_result_all}
	\begin{center}
		\begin{tabular}{lccc}
			\hline
			\multicolumn{1}{c}{\bf Variant} & \multicolumn{3}{c}{\bf Balanced accuracy} \\
			& Short & Middle & Long\\
			\hline \hline
            BaseCNN & 0.848~(0.012) & 0.917~(0.016) & \textbf{0.888}~(0.014)\\
            BaseCNN-te & 0.853~(0.008) & 0.919~(0.013) & 0.884~(0.015)\\
            \hline
            MSCNN & 0.884~(0.013) & 0.924~(0.015) & 0.869~(0.024)\\
            MSCNN-te & 0.885~(0.007) & 0.926~(0.016) & 0.868~(0.016)\\
            \hline
            ASCNN & 0.875~(0.011) & 0.915~(0.012) & 0.886~(0.021)\\
            ASCNN-te & 0.863~(0.012) & 0.919~(0.009) & 0.883~(0.017)\\
            \hline
            AMSCNN & 0.857~(0.011) & 0.910~(0.012) & 0.867~(0.016)\\
            AMSCNN-te & \textbf{0.895}~(0.015) & \textbf{0.928}~(0.014) & 0.867~(0.019)\\
			\hline
		\end{tabular}
	\end{center}
\end{table}


BaseCNN perform the best on long data, but the worst on short data even with TE. This may be because the feature extraction in BaseCNN is affected by the large padding region for short data.
AMSCNN performs the best on short data of all variants, although there are no significant differences between AMSCNN, MSCNN and ASCNN in their accuracies. MSCNN shows similar accuracies to AMSCNN and marks the second highest accuracy on short and middle data. ASCNN does not have the additional option to use local features with a small RF for long data, but this does not lead to a noticeable difference on long data in this experiment.


\begin{figure}[t]
	\begin{center}
        \includegraphics[width=7.5cm,trim=0 30 20 35,clip]{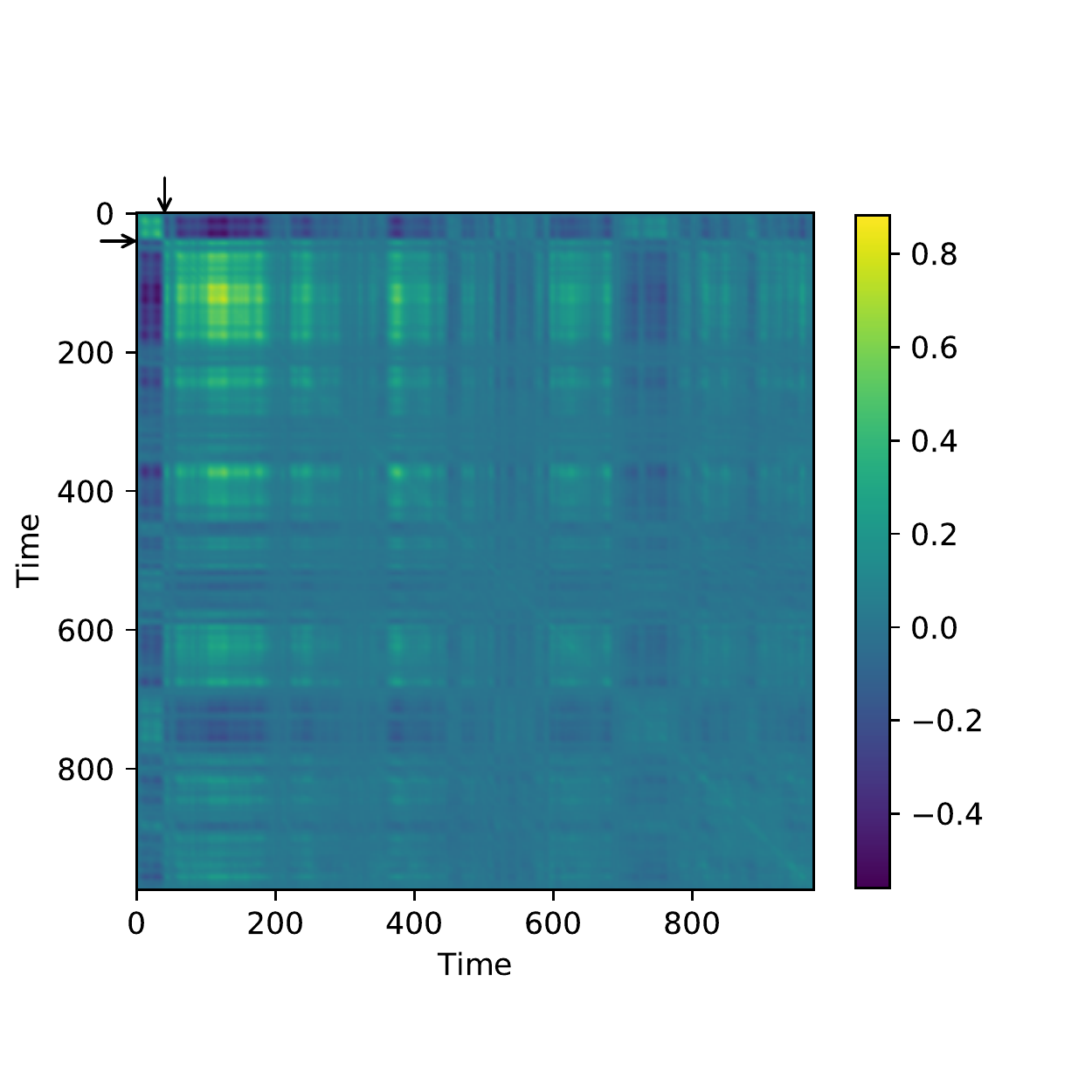}
	\end{center}
	\caption{The correlation matrices of TE extracted from AMSCNN-te (after the first convolution). Low values are shown in dark blue and high values are in yellow. Arrows point at time $40$.}
	\label{fig:TEmat}
\end{figure}

The correlation of each time component in TE is shown in Figure~\ref{fig:TEmat}. We find that the learned TE exhibits a structured pattern and has clear gradients at around $t=40$ (pointed by arrows), reflecting the dynamics when the liquid containers rotate.



\subsection{The UEA/UCR archive}\label{sec:exp1}
To investigate the effect and data-dependency of the two modules in detail, we conduct experiments on a subset of datasets in the UEA/UCR archive~\cite{UEA2018,UCR2019}. We omit datasets with less than 1,000 samples for training or test split in order to avoid the small sample effect. Tables~\ref{tab:ucr_data} and \ref{tab:uea_data} show the properties of the univariate and multivariate TSC datasets used in this experiment, respectively. Note that our method can be applied to various case of pTSC as simulated using this datasets, for example, the classification of corrupted time series, such as classification of choppy sounds and character recognition with disjoint strokes.

\begin{table}[t]
	\caption{Datasets from UCR univariate time series archive.}
	\label{tab:ucr_data}
	\begin{center}
		\begin{tabular}{lcccc}
			Name & Classes & Length & Train / Test \\
			\hline
			Crop & 24 & 46 & 7200 / 16800 \\
			ElectricDevices & 7 & 96 & 8926 / 7711 \\
			FordA & 2 & 500 & 3601 / 1320 \\
			MelbournPedestrian & 10 & 24 & 1200 / 2450 \\
			NonInvasiveFatalEGGThorax1 & 42 & 750 & 1800 / 1965 \\
			NonInvasiveFatalEGGThorax2 & 42 & 750 & 1800 / 1965 \\
			StarLightCurves & 3 & 1024 & 1000 / 8236 \\
			TwoPatterns & 4 & 128 & 1000 / 4000 \\
			Wafer & 2 & 152 & 1000 / 6164 \\
		\end{tabular}
	\end{center}
\end{table}

\begin{table}[t]
	\caption{Datasets from UEA multivariate time series archive.}
	\label{tab:uea_data}
	\begin{center}
		\begin{tabular}{lcccc}
			Name & Classes & Channels & Length & Train / Test \\
			\hline
			PhonemeSpectra & 39 & 11 & 217 & 3315 / 3353 \\
			SpokenArabicDigits & 10 & 13 & [4,93] & 6599 / 2199 \\
			CharacterTrajectories & 20 & 3 & [60,182] & 1422 / 1436 \\
			PenDigits & 10 & 2 & 8 & 7494 / 3498  \\
			LSST & 14 & 6 & 36 &  2459 / 2466  \\
			FaceDetection & 2 & 144 & 62 &  5890 / 3524  \\
			InsectWingbeat & 10 & 200 & [2,22] & 30000\,/\,20000  \\
		\end{tabular}
	\end{center}
\end{table}




We use ResNet~\cite{strongBaseline2016} as the base structure in this experiment rather than the state-of-the-art TSC models InceptionNet~\cite{inception2020} and OS-CNN~\cite{OS-CNN} because these two models have too wide RF sizes in a single convolutional block. The original ResNet has kernel sizes $8,5,3$, but we change them to odd numbers $9,5,3$ for symmetric padding. 
The most naive baseline is ResNet, which receives fixed-length inputs after preprocessing by interpolation. Another baseline is ResNet-vl, which accepts variable length inputs by controlling the target length in batch normalization and GAP. This is implemented by masking out the regions including padding values but keeping the length of at least $1$ even for inputs shorter than RF size. 
Models with the proposed components are based on ResNet-vl. ``AMSResNet" has AMSpool with $64$ channels for all additional layers instead of GAP. ``ResNet-te" and ``AMSResNet-te" have TE and the number of input channels is doubled in the first convolutional kernel. 
All models have a fixed and common RF size $43$ in this experiment on datasets with different lengths. 






The training settings are mostly chosen following the existing work~\cite{review2019}: the minibatch size is $16$, the maximum epoch is $1,000$ with early-stopping, the learning rate is initially $0.001$ and scheduled by multiplying $0.5$ when loss has not decreased for $50$ consecutive epochs down to the minimum value $0.0001$. The criterion for early-stopping and learning rate scheduling is validation loss evaluated on held-out $10\%$ data of the training set. In order to teach models to classify partial time series, we use random crop as data augmentation. The ratio of random crop is scheduled to ramp down linearly from $1.0$ (no crop) to $0.1$ until $800$ epochs because the models fail to converge with random crop in maximum ratio from the initial epoch. When calculating validation loss, we use the average over three fixed patterns: the original data, the first and latter half crop. All experiments are repeated $10$ times with different random seeds.
We report training and inference time for single minibatch with 16 data in Crop dataset evaluated on NVIDIA GTX 1080Ti embedded system using our PyTorch implementation. The training (inference) time is 15 (2.2) ms, 29 (7.8) ms and 35 (9.5) ms for ResNet, ResNet-vl and AMSRes-te, respectively. AMSRes-te needs additional time due to several layers in AMSpool compared with ResNet-vl. In our implementation, masking processes in GAP and batch normalization layers cause large increase in training and inference time compared with ResNet.

\begin{table}[t]
	\caption{Pairwise counts of wins and loses of AMSRes-te versus baseline ResNets (on half-cropped~/~complete data).}
	\label{tab:tsc_result_0}
	\begin{center}
		\begin{tabular}{lccc}
			Baseline & Win & Lose & Not significant \\
			\hline
			ResNet & 6~/~4 & 2~/~3 & 8~/~9 \\
			ResNet-vl & 4~/~4 & 3~/~2 & 9~/~10 \\
		\end{tabular}
	\end{center}
\end{table}

\begin{table*}[t]
	\caption{Results on UEA/UCR archive. Values are median accuracies in 10 runs (on half-cropped~/~complete data). Numbers in parentheses represent the standard deviation. The highest value in each row is in bold font. Underlines indicate values without significant difference from the highest one of the same row.}
	\label{tab:tsc_result}
	\begin{center}
		\begin{tabular}{lccccc}
			\hline
			\multicolumn{1}{c}{\bf Dataset} & \multicolumn{5}{c}{\bf Accuracy (on half-cropped / complete data)} \\
			& ResNet & ResNet-vl & ResNet-te & AMSResNet-vl & AMSResNet-te \\
			\hline \hline
			MelbournePedestrian & 0.431~/~\underline{0.742} & \textbf{0.562}~/~\underline{0.762} & \underline{0.515}~/~\underline{0.746} & 0.211~/~0.206 & \underline{0.534}~/~\textbf{0.790}\\
            Crop & 0.206~/~0.588 & \underline{0.530}~/~\underline{0.660} & \textbf{0.533}~/~\textbf{0.688} & 0.273~/~0.500 & 0.307~/~0.570\\
            ElectricDevices & 0.586~/~\underline{0.710} & \underline{0.644~}/~\underline{0.715} & \underline{0.645}~/~\underline{0.718} & \textbf{0.654}~/~\textbf{0.721} & \underline{0.652}~/~\underline{0.716}\\
            TwoPatterns & 0.515~/~\underline{0.962} & \underline{0.607}~/~0.938 & 0.571~/~\textbf{0.988} & \textbf{0.611}~/~0.931 & 0.572~/~\underline{0.987}\\
            Wafer & 0.958~/~\textbf{0.996} & 0.975~/~\underline{0.993} & \textbf{0.988}~/~\underline{0.995} & 0.944~/~\underline{0.989} & \underline{0.983}~/~\underline{0.994}\\
            FordA & \textbf{0.899}~/~\textbf{0.936} & \underline{0.894}~/~\underline{0.929} & 0.883~/~0.920 & \underline{0.895}~/~\underline{0.930} & 0.886~/~0.919\\
            NonInvasiveFetalECGThorax1 & \underline{0.069}~/~\underline{0.171} &\textbf{ 0.104}~/~\underline{0.164} & \underline{0.096}~/~\underline{0.141} & \underline{0.095}~/~\textbf{0.209} & \underline{0.066}~/~\underline{0.122}\\
            NonInvasiveFetalECGThorax2 & \underline{0.067}~/~\underline{0.160} & \underline{0.109}~/~\underline{0.202} & \underline{0.091}~/~\underline{0.187} & \textbf{0.110}~/~\textbf{0.263} & \underline{0.073}~/~\underline{0.155}\\
            StarLightCurves & \underline{0.862}~/~\underline{0.937} & \underline{0.874}~/~\underline{0.939} & \textbf{0.884}~/~\textbf{0.940} & \underline{0.882}~/~\underline{0.929} & \underline{0.779}~/~\underline{0.936}\\
            \hline
            InsectWingbeat & \textbf{0.580}~/~\textbf{0.664} & 0.432~/~0.583 & 0.445~/~0.591 & 0.480~/~0.622 & 0.500~/~0.643\\
            SpokenArabicDigits & \underline{0.672}~/~0.971 & \underline{0.838}~/~\underline{0.990} & \underline{0.807}~/~\textbf{0.991} & \underline{0.857}~/~\underline{0.988} & \textbf{0.866}~/~\underline{0.990}\\
            PenDigits & 0.459~/~0.847 & \textbf{0.784}~/~\underline{0.960} & \underline{0.742}~/~\underline{0.957} & 0.730~/~\underline{0.965} & \underline{0.748}~/~\textbf{0.969}\\
            LSST & 0.476~/~0.603 & 0.538~/~0.598 & 0.527~/~0.568 & \textbf{0.577}~/~\textbf{0.662} & \underline{0.573}~/~\textbf{0.662}\\
            CharacterTrajectories & 0.602~/~0.908 & 0.709~/~\underline{0.978} & 0.591~/~0.932 & \underline{0.783}~/~\underline{0.979} & \textbf{0.826}~/~\textbf{0.985}\\
            FaceDetection & 0.534~/~0.552 & 0.539~/~0.558 & \underline{0.604}~/~\underline{0.669} & 0.540~/~0.566 & \textbf{0.614}~/~\textbf{0.671}\\
            PhonemeSpectra & 0.160~/~\underline{0.268} & \textbf{0.232}~/~\textbf{0.271} & 0.160~/~0.206 & \underline{0.229}~/~\underline{0.261} & 0.128~/~0.172\\
			\hline
		\end{tabular}
	\end{center}
\end{table*}

\begin{figure}[t]
	\begin{center}
		\includegraphics[width=8.8cm,trim=20 240 100 100,clip]{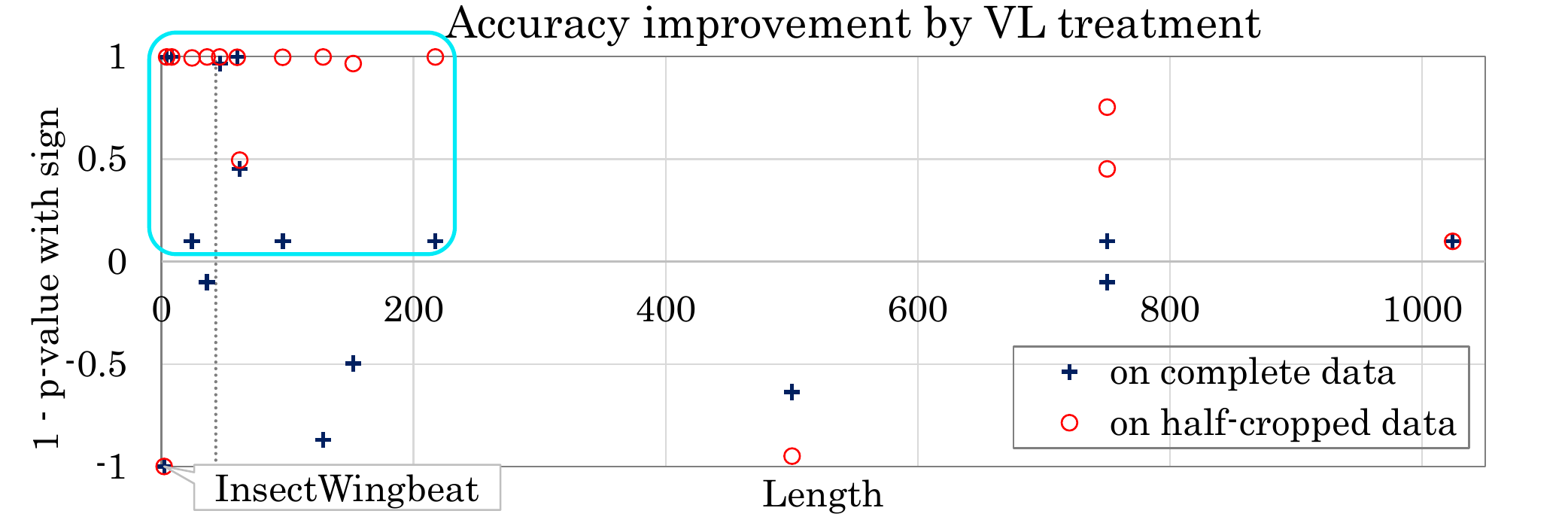}
	\end{center}
	\caption{Signed significance plot about variable-length (VL). Cyan rounded rectangle encloses the region where positive results are dominant. Dashed vertical line indicates length equals to the RF of ResNet.}
	\label{fig:VL_comp}
\end{figure}

\begin{figure}[t]
	\begin{center}
		\includegraphics[width=8.8cm,trim=20 240 100 100,clip]{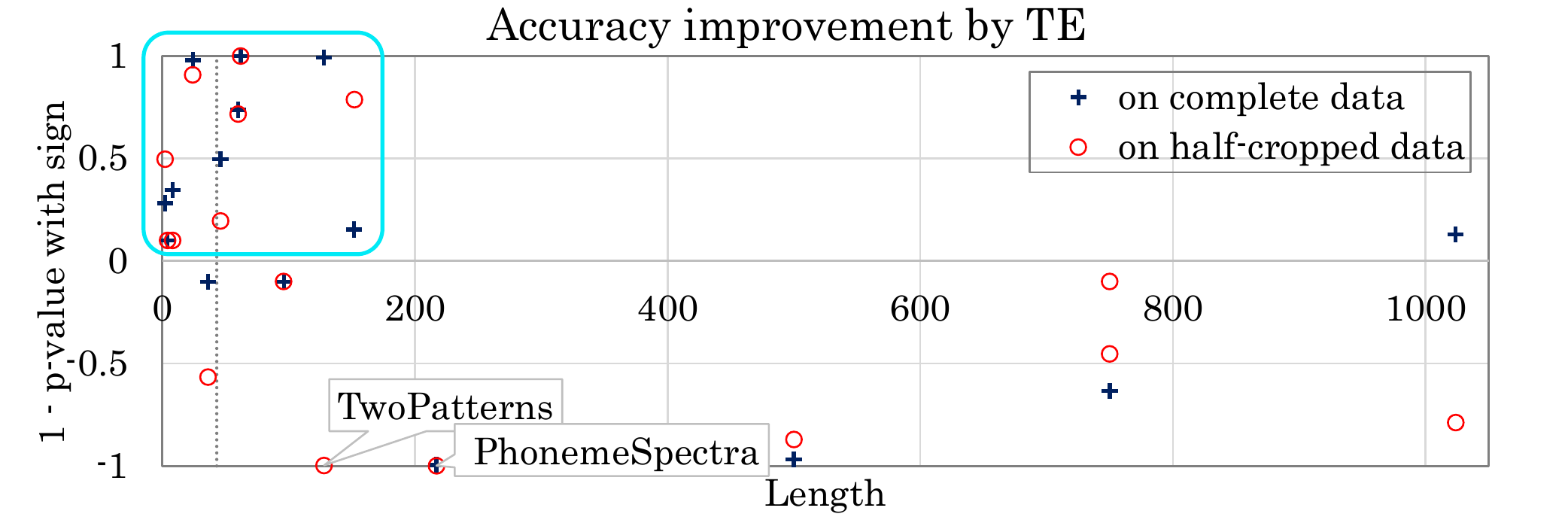}
	\end{center}
	\caption{Signed significance plot about TE. Cyan rounded rectangle encloses the region where positive results are dominant. Dashed vertical line indicates length equals to the RF of ResNet.}
	\label{fig:TE_comp}
\end{figure}

\begin{figure}[t]
	\begin{center}
		\includegraphics[width=8.8cm,trim=20 240 100 100,clip]{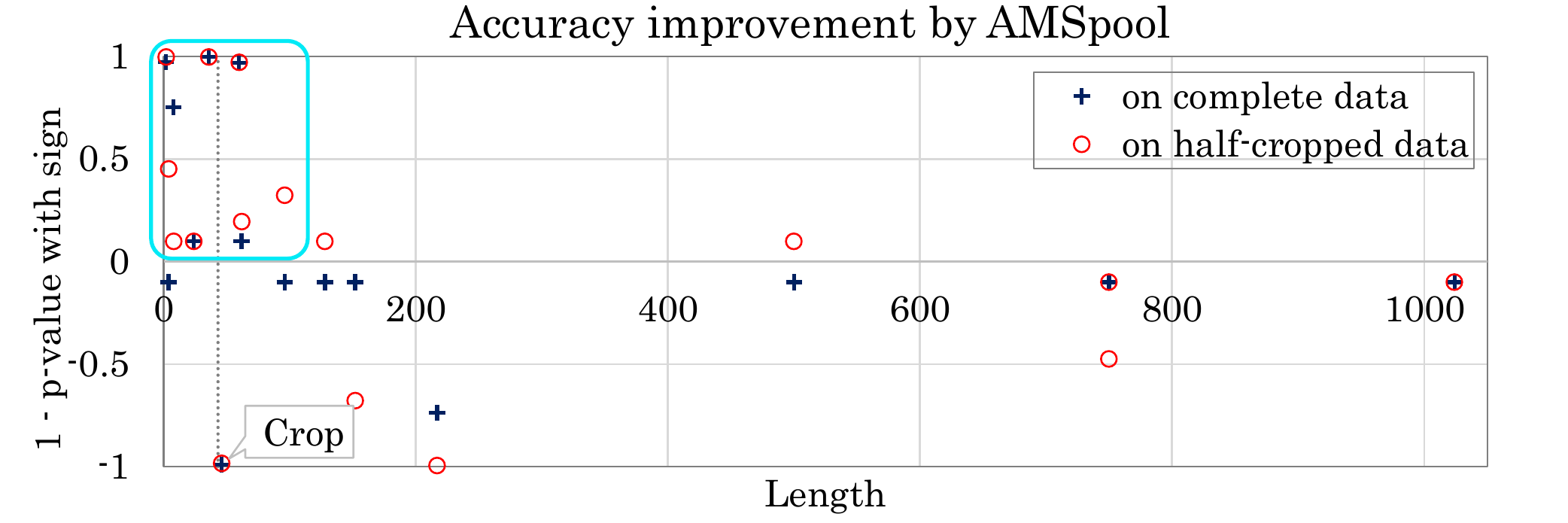}
	\end{center}
	\caption{Signed significance plot about AMSpool. Cyan rounded rectangle encloses the region where positive results are dominant. Dashed vertical line indicates length equals to the RF of ResNet.}
	\label{fig:AMSpool_comp}
\end{figure}

To compare the performance on pTSC, we evaluate the models on half-cropped data (using both the first and latter crop) as well as on the original test data. We conduct the statistical test Steel-Dwass method with p-value $0.05$ to clarify which datasets and pairs show significant differences. Table~\ref{tab:tsc_result_0} summarizes the result of the comparison between AMSResNet-te and two baselines ResNet and ResNet-vl. AMSResNet-te shows more wins than both of baselines, but there are non-negligible number of loses. Table~\ref{tab:tsc_result} shows the list of the accuracies on complete data and half-cropped data. The rankings of models differ between datasets, and we provide the discussion in the viewpoint of the dependency on length of datasets below.
Significance and sign of accuracy difference are summarized in Figs.~\ref{fig:VL_comp}, \ref{fig:TE_comp} and \ref{fig:AMSpool_comp} corresponding to the effect of variable-length treatment (ResNet-vl vs. ResNet), TE (AMSResNet-te vs. AMSResNet-vl) and AMSpool (AMSResNet-te vs. AMSResNet-vl), respectively. The plots show $1-$p-value with the sign of differences against the minimum length in each dataset to visualize length-dependency. Note that we cannot conclude there are significant difference if $|1-p|<0.95$.


As seen in Fig.~\ref{fig:VL_comp}, ResNet trained with random crop shows lower accuracy than ResNet-vl even for the original test data on some of datasets with length smaller than or equal to $60$. This degrade of accuracy occurs in almost all datasets with length smaller than or equal to $217$ on half-cropped data. This is because time-scale information for the time series is destroyed by inconsistent interpolation in preprocessing. Only InsectWingbeat dataset shows negative effect of variable length models. We guess this is because the length of this dataset is too short (the minimum length equals $2$) and the feature extracted by ResNet-vl is smeared out by padding value.
From Fig.~\ref{fig:TE_comp}, positive effect of TE can be found on length smaller than $217$ and causes negative effect on PhonemeSpectra. On TwoPatterns at length $128$, the effect is reversed between the test on complete data and on half-cropped data. This is contradictory to our expectation that TE can compensate time information and contribute to accuracy especially on short data. 

The significant improvements with AMSpool are found on datasets whose lengths are comparable to or shorter than the RF of ResNet as shown in Fig.~\ref{fig:AMSpool_comp}.
This is because AMSpool can adjust RF to handle short inputs within the RF of a base CNN. Crop, which has the closest length to RF, is an exception that shows a negative difference due to AMSpool. We guess the reason is that the models with AMSpool cannot exploit last convolutional block when almost all train data are given after cropped by length shorter than the RF. We provide the additional experiments using Crop and LSST (one of datasets which shows positive difference) with length extended to $1.25$ times as long as the original by interpolation. However, it shows still a negative difference on Crop and a positive difference on LSST by AMSpool. This suggests the cause of the exception is not explained by length of datasets.

\section{Conclusion}\label{sec:conclusion}
We defined the novel challenging problem pTSC, where lengths and timestamps of inputs vary.
To address the issue, we proposed two modules AMSpool and TE.
AMSpool adjusts the receptive field (RF) by aggregating features from adaptive number of layers with different RFs. This enables us to resolve the trade-off between short data and long data in classification using CNN. 
TE concatenates the embedding of timestamps to inputs as additional channels to tell the difference of timestamps.

Experiments on our trajectory dataset have shown that our modules improve classification accuracy especially on short data. The other experiments using the UEA/UCR archive have shown that AMSpool improves classification accuracy on the datasets with the length comparable to the RF of the final convolutional features.

In the former experiment, accuracy on long data is a little lower than the fixed-length CNN used as the base structure. To solve this degradation is left for future work.



\bibliographystyle{IEEEtran}
\bibliography{ref_tsc}

\end{document}